\documentclass[11pt,a4paper]{article}
\usepackage[hyperref]{acl2018}
\usepackage{times}
\usepackage{latexsym}
\usepackage{graphicx}

\usepackage{url}

\let\ACMmaketitle=\maketitle
\renewcommand{\maketitle}{\begingroup\let\footnote=\thanks \ACMmaketitle\endgroup}

\aclfinalcopy % Uncomment this line for the final submission
\def\aclpaperid{15} %  Enter the acl Paper ID here

\title{Towards Automation of Sense-type Identification of Verbs in OntoSenseNet(Telugu) \footnote{ * This work was presented at $6^{th}$ International Workshop on Natural Language Processing for Social Media (SocialNLP) at $56^{th}$ Annual Meeting of the Association for Computational Linguistics, ACL.} }

\author{Sreekavitha Parupalli, Vijjini Anvesh Rao and Radhika Mamidi\\
  Language Technologies Research Center (LTRC) \\
  International Institute of Information Technology, Hyderabad \\
  {\tt \{sreekavitha.parupalli, vijjinianvesh.rao\}@research.iiit.ac.in } \\
  {\tt radhika.mamidi@iiit.ac.in} \\}

\date{}

\begin{document}
\maketitle
\begin{abstract}
In this paper, we discuss the enrichment of a manually developed resource of Telugu lexicon, OntoSenseNet. OntoSenseNet is a ontological sense annotated lexicon that marks each verb of Telugu with a primary and a secondary sense. The area of research is relatively recent but has a large scope of development. We provide an introductory work to enrich the OntoSenseNet to promote further research in Telugu. Classifiers are adopted to learn the sense relevant features of the words in the resource and also to automate the tagging of sense-types for verbs. We perform a comparative analysis of different classifiers applied on OntoSenseNet. The results of the experiment prove that automated enrichment of the resource is effective using SVM classifiers and Adaboost ensemble.
\end{abstract}

\section{Introduction}
\label{intro}

Telugu is morphologically rich and follows different grammatical structures compared to western languages such as English and Spanish. However, to maintain compatibility, the western ideology of rules are adopted in current approaches. Thus, many ideas and significant information of the language is lost. Indian languages are generally fusional (Hindi, English) and agglutinative in nature (Telugu) \cite{pingali2006hindi}. The morphological structure of agglutinative language is unique and capturing its complexity in a machine analyzable and reproducible format is a challenging job \cite{dhanalakshmi2009morphological}. 

OntoSenseNet is a lexical resource developed on the basis of Formal Ontology proposed by \cite{otra2015towards}. The formal ontology follows approaches developed by Yaska, Patanjali and Bhartrihari from Indian linguistic traditions for understanding lexical meaning and by extending approaches developed by Leibniz and Brentano in the modern times. This framework proposes that meaning of words are in-formed by intrinsic and extrinsic ontological structures \cite{rajan2015ontological}. 

Based on this proposed formal ontology, a lexical resource for Telugu language has been developed \cite{2018arXiv180402186P}. The resource consists of words tagged with a primary and a secondary sense. The sense-identification in OntoSenseNet for Telugu is done manually by experts in the field. But, further manual annotation of the immense amount of corpus proves to be cost-ineffective and laborious. Hence, we propose a classifier based automated approach to further enrich the resource. The fundamental aim of this paper is to validate and study the possibility of utilizing machine learning algorithms in the task of automated sense-identification.

\section{Related Work}
\label{sec:relatedwork}
The work contributes to building a strong foundation of datasets in Telugu language to enable further research in the field. This section describes previously compiled datasets available for Telugu and past work related to our dataset. We also talk about some recent advancements in NLP tasks on Telugu.

Telugu WordNet, developed as part of IndoWordNet\footnote{\url{http://www.cfilt.iitb.ac.in/indowordnet/index.jsp}}, is an exhaustive set of multilingual assets of Indian languages. Telugu WordNet is introduced to capture semantic word relations including but not limited to hypernymy-hyponymy and synonymy-antonymy.  

Recent advances are observed in several NLP tasks on Telugu language. \cite{choudhary2018emotions} developed a siamese network based architecture for sentiment analysis of Telugu and \cite{singh2018automatic} utilize a clustering-based approach to handle word variations and morphology in Telugu. But, the ideology that forms the basis of their assumptions lies in western ideology inspired from major western languages. This is due to lack of a large publicly available resource based on the ideology of senses.

\section{Data Description}
\label{sec:dataset}

Telugu is a Dravidian language native to India. It stands alongside Hindi, English and Bengali as one of the few languages with official primary language status in India\footnote{\url{https://en.wikipedia.org/wiki/Telugu_language}}. Telugu language ranks third in the population with number of native speakers in India (74 million, 2001 census)\footnote{\url{https://web.archive.org/web/20131029190612/http://censusindia.gov.in/Census_Data_2001/Census_Data_Online/Language/Statement1.htm}}. However, the amount of annotated resources available is considerably low. This deters the novelty of research possible in the language. Additionally, the properties of Telugu are significantly different compared to major languages such as English. 

In this paper, we adopt the lexical resource OntoSenseNet for Telugu. The resource consists of 21,000 root words alongside their meanings. The primary and secondary sense of each extracted word is identified manually by the native speakers of language. The paper tries to automate the process and enrich the existing resource. The sense-type classification has been explained below in section \ref{sec:classification} . 

The dataset on which we trained the skip gram model \cite{mikolov2013distributed} consists of 27 million words extracted from Telugu Wikipedia dump. Further, we populated our dataset by adding 46,972 sentences from SentiRaama corpus\footnote{\url{https://ltrc.iiit.ac.in/showfile.php?filename=downloads/sentiraama/}} obtained from Language Technologies Research Centre, KCIS, IIIT Hyderabad. Additionally, we added 5410 lines obtained from \cite{mukku2016enhanced}. The corpus that has been assembled is one the of few datasets available in Telugu for research purpose. 

\subsection{Morphological Segmentation}
Telugu, being agglutinative language, has a high rate of affixes or morphemes per word. Thus, OntoSenseNet resource has little coverage over the Wikipedia data utilized to develop the vector space model. Hence, we applied morphological analysis on both OntoSenseNet and Wikipedia data to segment complex words into its subparts. This leads to an improvement in the coverage of OntoSenseNet resource over the dataset. Thus, the frequency of OntoSenseNet resource increases significantly in the wikipedia corpus. However, the problem of imbalanced class distribution still persists. The addition of this module is empirically justified by the improvements in over-all accuracy metrics shown in the evaluation of results (Section \ref{sec:results}).

\subsection{Sense-type classification of Verbs}
\label{sec:classification}
Verbs provide relational and semantic framework for its sentences and are considered as the most important lexical and syntactic category of language. In a single verb many verbal sense-types are present and different verbs share same verbal sense-types. These sense-types are inspired from different schools of Indian philosophies \cite{rajan2013understanding}. 
The seven sense-types of verbs along with their primitive sense along with Telugu examples are given by \cite{2018arXiv180402186P}. In this paper, we adopt 8483 verbs of OntoSenseNet as our gold-standard annotated resource. This resource is utilized for learning the sense-identification by classifiers developed in our paper. 

\begin{itemize}
\item Know|Known - To know.
Examples: \textit{daryāptu (investigate), vivaran̄a (explain)}

\item Means|End -  To do. 
Examples: \textit{parugettu (run), moyu (carry)}

\item Before|After - To move. 
Examples: \textit{pravāhaṁ (flow), oragupovu (lean)}

\item Grip|Grasp -  To have. 
Examples: \textit{lāgu (grab), vārasatvaṅga (inherit)}

\item Locus|Located - To be. 
Examples: \textit{Ādhārapaḍi (depend), kaṅgāru (confuse)}

\item Part|Whole - To cut.
Examples: \textit{perugu (grow), abhivṛddhi (develop)}

\item Wrap|Wrapped - To bound. 
Examples: \textit{dharin̄caḍaṁ (wear), Āśrayaṁ(shelter)}

\end{itemize}

% \begin{itemize}
% \item Know|Known - Conceptualize, construct or transfer information between or within an animal (To know). 

% \item Means|End - A process which cannot be accomplished without a doer (To do). 

% \item Before|After - Every process has a movement in it. The movement maybe a change of state or location (To move). 

% \item Grip|Grasp -  Possessing, obtaining or transferring a quality or object (To have). 

% \item Locus|Located - Continuously having (to be in a state) or possessing a quality (To be). 

% \item Part|Whole - Separation of a part from whole or joining of parts into a whole. Processes which causes a pain. Processes which disrupt the normal state (To cut).

% \item Wrap|Wrapped - Processes which pertain to a certain specific object or category. It is like a bounding (To bound). 

% \end{itemize}

\section{Methodology \& Training}
\label{sec:training}
We train a Word2Vec skip-gram model on 2.36 million lines of Telugu text. We train classifiers in one-vs-all setting to get prediction accuracy for each label. Furthermore, we trained and validated on the OntoSenseNet corpus explained in the previous section.

\begin{figure}[h]
\includegraphics[width=\columnwidth]{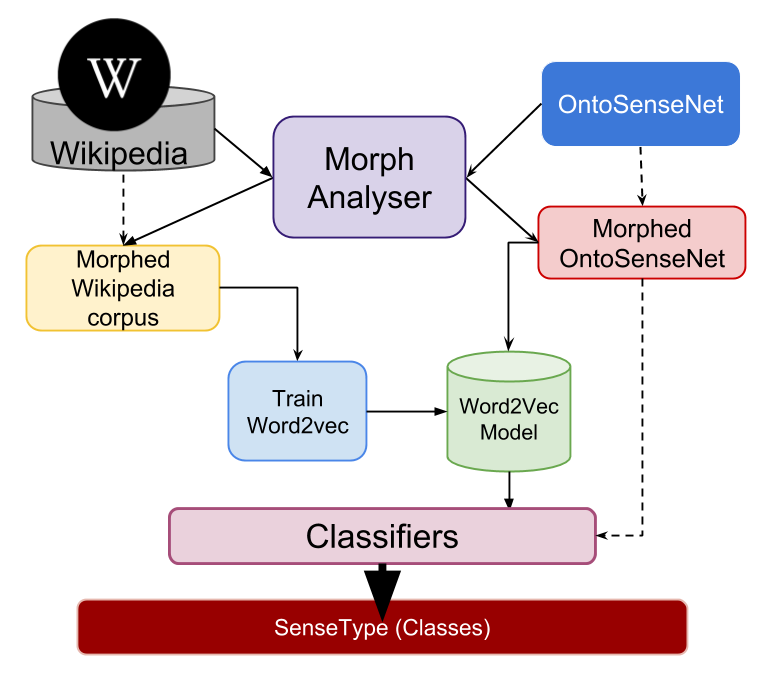}
\caption{Methodology}
\label{fig:system}
\end{figure}

\subsection{Pre-Processing and Training}

Figure \ref{fig:system} depicts the pre-processing steps and overall architecture of our system. To train the vector space embedding (Word2Vec), we initiate by deleting unwanted symbols, punctuation marks, especially ones that do not add significant information. After that, we perform the morphological segmentation of the data and split all the Telugu words in the large Word2Vec training corpus into individual morphemes. For this task, we utilize the Indic NLP library \footnote{\url{http://anoopkunchukuttan.github.io/indic_nlp_library/}} 
which provides morphological segmentation among other tools, for several Indian languages. Along with splitting morphemes to train Word2Vec, we also stem the words of OntoSenseNet resource. This process of morphological segmentation produces a significant rise in frequencies of morphemes, hence, promoting better vector representations for the Word2Vec model.   

Additionally, we only accept embeddings of words present in the OntoSenseNet resource for which an embedding exists in our trained Word2Vec model. This enables us to reduce the problem of resource enrichment to a classification task. To train the classifiers, we need the word embeddings of the OntoSenseNet's words. However, the words in the resource are also complex and agglutinative in nature. Hence, we stem the OntoSenseNet words too to the smallest root, so that we are able to search them with the Word2Vec embedding model. Finally, the morphed data of embedding training dataset is utilized for training Word2Vec, and stemmed OntoSenseNet words' vectors are extracted to train classifiers described in the next section (Section \ref{sec:classification}). We have used only primary sense-type tagging of the words in OntoSenseNet for enrichment. 
% ^^^^^^^^^^^^^^^^^^^^^^^^^^^^^^^^^^^

% We study and analyze several classifier approaches to choose the one with best results. The variants we considered are K Nearest Neighbors, Linear SVM, Gaussian SVM, Adaboost Ensemble, Decision Trees, Random Forest and Neural Networks. 

\subsection{Classifier based Approaches}
As each word can have any of the seven sense-types, we have a multi-class classification problem at hand. In Table {\ref{tab: sensetype}} , we show the multi-class classification accuracies for different classifiers. Additionally, in Figure {\ref{fig:accuracy chart}} and Figure {\ref{fig:quantitative chart}} we show the one-vs-all accuracies for the seven sense-types of verbs across different classifiers. 
We then study and analyze these classifier approaches to choose the one with best results. The variants we considered are discussed below:

\subsubsection{K Nearest Neighbors} 

K nearest neighbors is a simple algorithm which stores all available samples and classifies new sample based on a similarity measure (inverse distance functions). A sample  is classified by a majority vote of its neighbors, with the sample being assigned to the class most common amongst its K nearest neighbors measured by a distance function. 
% \footnote{\url{}}

\subsubsection{Support Vector Machines (SVM)}
SVM classifier is a supervised learning model that constructs a set of hyperplanes in a high-dimensional space which separates the data into classes. SVM is a non-probabilistic linear classifier. SVM takes the input data and for each input data row it predicts the class to which this input row belongs. 

The Gaussian kernel computed with a support vector is an exponentially decaying function in the input feature space, the maximum value of which is attained at the support vector and which decays uniformly in all directions around the support vector, leading to hyper-spherical contours of the kernel function. 

\subsubsection{Adaboost Ensemble}
An AdaBoost classifier is a meta-estimator that begins by fitting a classifier on the original dataset and then fits additional copies of the classifier on the same dataset but where the weights of incorrectly classified instances are adjusted such that subsequent classifiers focus more on difficult cases.

\subsubsection{Decision Trees} 
Decision tree (DT) can be described as a decision support tool that uses a tree like model for the decisions and their likely outcomes. A decision tree is a tree in which each internal (non-leaf) node is labeled with an input feature. Class label is given to each leaf of the tree. But for our work, decision tree gives less accurate results because of over-fitting on the training data. We took the tree depth as 5 for each decision tree.

\subsubsection{Random Forest} 
A Random Forest (RF) classifier is an ensemble of Decision Trees. Random Forests construct several decision trees and take each of their scores into consideration for giving the final output. Decision Trees have a great tendency to overfit on any given data. Thus, they give good results for training data but bad on testing data. Random Forests reduces over-fitting as multiple decision trees are involved. We took the n estimator parameter as 10.

\subsubsection{Neural Networks} 
Multi layer perceptron (MLP) is a feedforward neural network with one or more layers between input and output layer. We call it feedforward as the data flows from input to output layer in a forward manner. Back propagation learning algorithm is used in the training for this sort of network. Multi layer perceptron is found very useful to solve problems which are not linearly separable. The neural network we use for our problem  has two hidden layers with the respective sizes being 100 and 25. 

\section{Evaluation of the Results}
\label{sec:results}

We have performed qualitative and quantitative analysis on the results obtained to study the aforementioned experiments. 

\begin{figure*}[h]
\includegraphics[width=\textwidth]{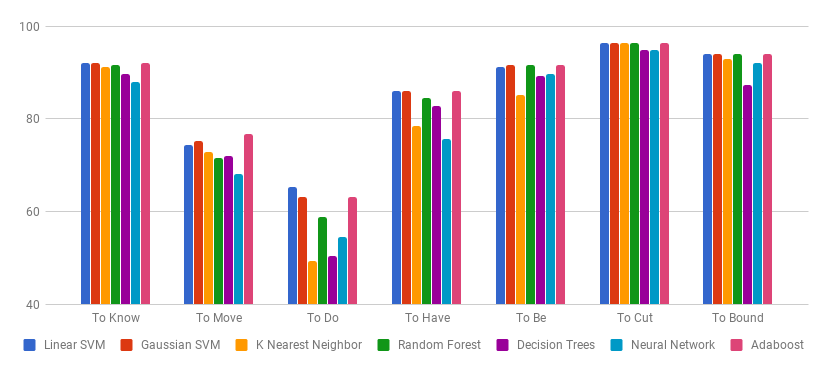}
\caption{Accuracies for all the sense-types of verbs when the classifiers are trained in one-vs.-all setting. }
\label{fig:accuracy chart}
\end{figure*}

%%%%%%%%%%TABLE%%%%%%%%%
\begin{table}
% \begin{center}
\begin{tabular}{| c | c  c |}
\hline
\textbf{Classifiers}&\textbf{Before}&\textbf{After}\\
\hline
Linear SVM & 35.34\% &40.72\% \\
Gaussian SVM & 36.78\% & 42.05\% \\
K Nearest Neighbor & 26.82\% & 27.48\% \\
Random Forest & 33.76\% & 37.08\% \\
Decision Trees & 33.50\% & 35.09\% \\
Neural Network & 31.67\% & 40.39\% \\
Adaboost & 34.43\% & 34.68\% \\
\hline
\end{tabular}
% \end{center}
\caption{Improvement of over-all classification accuracy \textit{before} and \textit{after} Morphological Segmentation.}
\label{tab: sensetype}
\end{table}

%%%%%%%%%%%%%%%%%%%%%

\subsection{Qualitative Analysis}

The results (depicted in Figure \ref{fig:accuracy chart}) portray that certain sense-types are predicted with significantly better accuracy than others. The experiments on ``To Do'' sense-type, especially, result in low accuracy relative to the other sense-types. In the resource, number of samples in one sense-type is higher than others, leaving other sense-types with fewer examples. Furthermore, different types of classifiers produce approximately similar accuracies in identifying particular sense-types. This is due to poor coverage of OntoSenseNet resource in the chosen corpus and also due to difference in distribution of sense-types in the Telugu language. However, we train the classifiers on equal distribution of the sense-types. But, the validation covers the entire OntoSenseNet. Thus, the imbalance in the sense-type distribution of the OntoSenseNet results in low accuracies for the sense-types with more number of samples in the validation set (including ``To do'').

Additionally, we justify the addition of morphological analyzer due to its added performance boost of over-all accuracy (shown in Table \ref{tab: sensetype}). 

Furthermore, of the 21,000 root words present in the OntoSenseNet database, only a one-third of the resource have embeddings present in the Word2Vec model, even after stemming. One of the major reasons is that the first volume of the current de facto dictionary was developed in 1936. Language dialects undergo critical evolution with influence from several languages such as Hindi, Tamil and English over time. The corpus adopted in the paper for training the vector space model mainly consists of Telugu Wikipedia data along with some recent collections of various online Telugu News, Books and Poems, that was created relatively recently (in the last decade). 

Figure \ref{fig:accuracy chart} displays that while the relative difference among classifiers is less as compared to performance across sense types, there are still some performance patterns that are observed. Across majority of the metrics, Gaussian SVM performs the best and outperforms all the classifiers including linear SVM indicating that the data is linearly separable in higher dimensions. Another commonly noted observation is that of Decision Tree versus Random Forest. Decision Trees tend to perform worse than Random Forest as they overfit on large data. However, Random Forests circumvent this problem by having multiple or an ensemble of decision trees, leading to a better performance, which is also reflected in our experiments. 

% k-NN shows relatively low performance which, also, supports our inference. 

%%%%FIGURE%%%%%%%%%%%%%%%
\begin{figure}[h]
\includegraphics[width=\columnwidth]{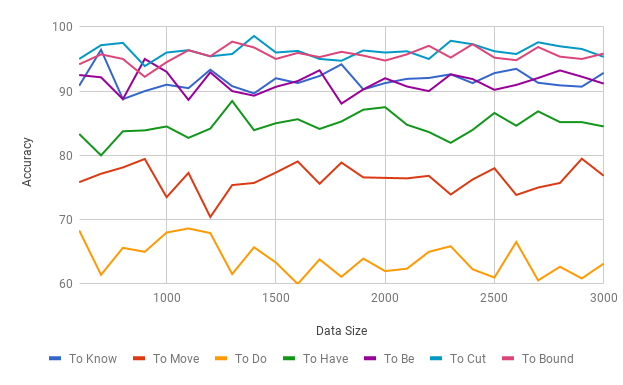}
\caption{Accuracy of each sense-type across changing number of data samples using a Gaussian SVM.}
\label{fig:quantitative chart}
\end{figure}
%%%%%%%%%%%%%%%%%%%%%%%%%%

\subsection{Quantitative}
For quantitative analysis, to understand the correlation between accuracy performance and training size, we choose Gaussian SVM as the classifier because it gives the best results (Figure \ref{fig:accuracy chart}). The graph of accuracy of each sense-type, given the classifier is a Gaussian SVM, is illustrated in Figure \ref{fig:quantitative chart}. A major observation from the results is the consequence of class imbalance. The initial increase in data results in a boost in performance of the model. But, as the number of samples in the test data increases, the class imbalance of the validation dataset becomes more prominent leading to fluctuations in the accuracy.

\section{Conclusion and Future Work}
\label{sec:conclusion}

Automatic enrichment of OntoSenseNet is attempted in this work. We compare several classifiers and test, validate their effectiveness in the task. Qualitative analysis of the classifiers empirically proves that Gaussian SVM is the best for the task of enriching OntoSenseNet. Quantitative analysis proves that, given a method to handle class imbalance, the model's effectiveness is directly proportional to the amount of training data. A continuation to this paper could be handling adjectives and adverbs available in OntoSenseNet for Telugu. Additionally, we identify a case of clustering-based extension like fuzzy k means where each word has a probability of belonging to each sense-type, rather than completely belonging to just one. This helps in identification of the secondary senses of verbs in OntoSenseNet.

\subsection{Acknowledgments}
We would like to thank Nurendra Choudary for helping us in formulation and development of this idea.

%\bibliography{acl2018}
\bibliography{acl2018}
\bibliographystyle{acl_natbib}

\end{document}